\pdfoutput=1
\documentclass[11pt]{article}

\usepackage{EMNLP2022}

\usepackage{times}
\usepackage{latexsym}

\usepackage[T1]{fontenc}
\usepackage[utf8]{inputenc}

\usepackage{microtype}

\usepackage{inconsolata}
\usepackage{times}
\usepackage{latexsym}
\usepackage{enumitem}
\usepackage[T1]{fontenc}
\usepackage[utf8]{inputenc}

\usepackage{xspace}
\usepackage{todonotes}
\usepackage{booktabs}
\usepackage{soul}
\usepackage{algorithm}
\usepackage{algorithmic}
\usepackage{newfloat}
\usepackage{listings}

\usepackage{booktabs}
\usepackage{graphicx}
\usepackage{subcaption} % For subtables

\newcommand{\iclaim}{\emph{Input}\xspace}
\newcommand{\vclaim}{\emph{Verified}\xspace}
\newcommand{\politifact}{PolitiFact\xspace}

\newcommand{\vclaimstatement}{\emph{VerifiedStatement}\xspace}
\newcommand{\vclaimtitle}{\emph{Title}\xspace}
\newcommand{\vclaimtext}{\emph{Body}\xspace}

\newcommand{\agree}{\emph{agree}\xspace}
\newcommand{\disagree}{\emph{disagree}\xspace}

\title{Assisting the Human Fact-Checkers:\\ Detecting All Previously Fact-Checked Claims in a Document}

\author{
    Shaden Shaar$^1$\thanks{\hspace{1.5mm}Work done while author was at QCRI, HBKU.},
    Nikola Georgiev$^2$,
    Firoj Alam$^3$,\\ 
    \bf Giovanni Da San Martino$^4$, 
    \textbf{Aisha Mohamed$^5$, Preslav Nakov$^6$}\\
  $^1$Cornell University,
  $^2$Sofia University,
  $^3$Qatar Computing Research Institute, HBKU, \\
  $^4$University of Padova,
  $^5$University of Wisconsin-Madison,\\
  $^6$Mohamed bin Zayed University of Artificial Intelligence\\
  \texttt{sshaar31@gmail.com, nikikg95@gmail.com, falam@hbku.edu.qa}\\
  \texttt{dasan@math.unipd.it, ahmohamed2@wisc.edu, preslav.nakov@mbzuai.ac.ae} 
  \\\\
  }

\begin{document}
\maketitle
\begin{abstract}
Given the recent proliferation of false claims online, there has been a lot of manual fact-checking effort. As this is very time-consuming, human fact-checkers can benefit from tools that can support them and make them more efficient. Here, we focus on building a system that could provide such support. Given an input document, it aims to detect all sentences that contain a claim that can be verified by some previously fact-checked claims (from a given database). The output is a re-ranked list of the document sentences, so that those that can be verified are ranked as high as possible, together with corresponding evidence. Unlike previous work, which has looked into claim retrieval, here we take a document-level perspective. We create a new manually annotated dataset for this task, and we propose suitable evaluation measures. We further experiment with a learning-to-rank approach, achieving sizable performance gains over several strong baselines. Our analysis demonstrates the importance of modeling text similarity and stance, while also taking into account the veracity of the retrieved previously fact-checked claims. We believe that this research would be of interest to fact-checkers, journalists, media, and regulatory authorities.
\end{abstract}

\section{Introduction}

\begin{figure}[tbh]
\centering
\small
\includegraphics[width=0.42\textwidth]{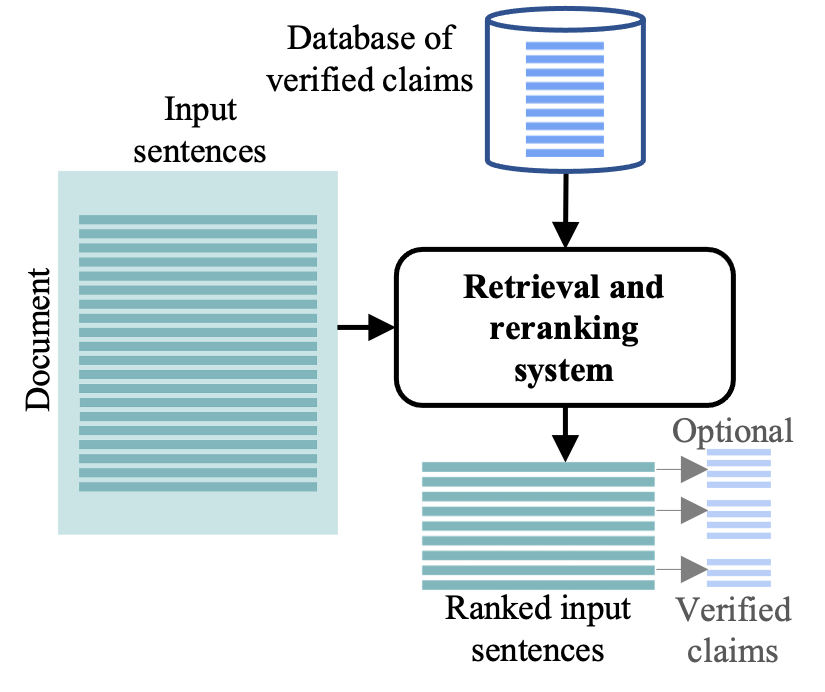}
\caption{The architecture of our system. Given an input document, it aims to detect all sentences that contain a claim that can be verified by some previously fact-checked claims (from a given database). The output is a re-ranked list of the document sentences, so that those that can be verified are ranked as high as possible, together with corresponding evidence.}
\label{fig:system_pipeline}
\end{figure}

Recent years have brought us a proliferation of false claims, which spread fast online, especially in social media; in fact, much faster than the truth~\cite{Vosoughi1146}. To deal with the problem, a number of fact-checking initiatives have been launched, such as FactCheck, FullFact, PolitiFact, and Snopes, where professional fact-checkers verify claims. Yet, manual fact-checking is very time-consuming and tedious.

%As checking a single claim can take many hours, even days~\cite{vlachos-riedel-2014-fact}, 
Thus, automatic fact-checking has been proposed as an alternative \cite{Li:2016:STD:2897350.2897352,Shu:2017:FND:3137597.3137600,rashkin-etal-2017-truth,Hassan:2017:CFE:3137765.3137815,vo2018rise,lee-etal-2018-improving,li-etal-2018-end,thorne-vlachos:2018:C18-1,Lazer1094,Vosoughi1146,zhang-etal-2020-answerfact,alam-etal-2022-survey,10.1145/3517214}. 
%It is indeed useful in many scenarios, 
While it scales better and works faster, it lags behind in quality, credibility, transparency, and explainability.
%, and it cannot rival the quality that manual fact-checking can offer. 

Manual and automatic fact-checking 
%will co-exist in the near future, and they 
can benefit from each other as automatic methods are trained on data that human fact-checkers produce, while human fact-checkers can be assisted by automatic tools. A middle ground between manual and automatic fact-checking is to verify an input claim by finding a previously fact-checked claim that allows us to make a true/false judgment on the veracity of the input claim. This is the problem we will explore below.

%the task of detecting mentions of previously fact-checked claims in a new text, e.g.,~in a political debate or speech, in a news report on a TV channel, in a tweet, etc. This is the problem we will explore below.

Previous work has approached the problem at the sentence level: given an input \emph{sentence/tweet}, produce a ranked list of relevant previously fact-checked claims that can verify it \cite{shaar-etal-2020-known}.
However, this formulation does not factor in whether the factuality of the input \emph{sentence/tweet} can be determined using the database of previously fact-checked claims, as it is formulated as a ranking task. For example, in a US presidential debate that has 1,300 sentences on average, only a small fraction would be verifiable using previously fact-checked claims from \politifact. Therefore, we target a more challenging reformulation at the \emph{document} level, where the system needs to prioritize which sentences are most likely to be verifiable using the database of previously fact-checked claims. This is still a ranking formulation, but here we rank the sentences in the input document (by verifiability using the database of claims), as opposed to ranking database claims for one input sentence (by similarity with respect to that sentence).
% However, here we target a more challenging reformulation at the \emph{document} level, where the system needs to prioritize the sentences to bring to the user's attention, by producing an appropriate ranking. For example, a US presidential debate has 1,300 sentences on average, but only a small fraction of them would be verifiable using \sh{a certain database of} previously fact-checked claims.
%Moreover, we are interested in returning actual matches, but not just such that agree or disagree with the input sentence, but ones that would guarantee that there would be no need to fact-check the input claim manually once we have found a good match. 

%Here, we focus on building a system that can help fact-checkers, journalists, and regulatory authorities when they need to analyze an entire document, e.g.,~a political debate, a news report on a TV channel, etc. 
In our problem formulation, given an input \emph{document}, the system needs to detect all sentences that contain a claim that can be verified by a previously fact-checked claim (from a given database of such claims).  The output is a re-ranked list of the document sentences, so that those that can be verified are ranked as high as possible, as illustrated in Figure~\ref{fig:system_pipeline}. The system could optionally further provide a corresponding fact-checked claim (or a list of such claims) from the database as evidence.
% The system could optionally further provide a corresponding claim (or a list of claims) from the database as evidence.
%Given a \emph{document}, we aim to (\emph{i})~find \emph{all sentences} in the document that contain claims that have been fact-checked before, and (\emph{ii})~return a list of corresponding matches from a database of previously fact-checked claims. 
Note that we are interested in returning claims that would not just be relevant when fact-checking the claims in the input sentence, but also would be enough to decide on a verdict for its factuality.
% on its factuality.

This novel formulation of the problem would be of interest to fact-checkers not only when they are facing a new document to analyze, but also when they want to check whether politicians keep repeating claims that have been previously debunked, so that they can be approached for comments.
It would also be of interest to journalists, as it could bring them a tool that can allow them to put politicians and public officials on the spot, e.g.,~during a political debate, a press conference, or an interview, by showing the journalist in real time which claims have been previously fact-checked and found false. Finally, media outlets would benefit from such tools for self monitoring and quality assurance, and so would regulatory authorities such as Ofcom.
%$\footnote{\url{http://www.ofcom.org.uk/}} 

Our contributions can be summarized as follows:
\begin{itemize}
%[noitemsep,topsep=0pt,leftmargin=10pt,labelwidth=!,labelsep=.5em]
%[leftmargin=*,noitemsep]
    \item We introduce a new real-world task formulation to assist fact-checkers, journalists, media, and regulators in finding which claims in a document have been previously fact-checked.
    \item We develop a new dataset for this task formulation, which consists of seven debates, 5,054 sentences, 16,636 target verified claims to match against, and 75,810 manually annotated sentence--verified claim pairs.
    \item We define new evaluation measures (variants of MAP), which are specifically tailored for our task. % setup.
    \item We address the problem using a learning-to-rank approach, and we demonstrate sizable performance gains over strong baselines.
    \item We offer analysis and discussion, which can facilitate future research.
    \item We release our data and code.\footnote{\url{https://github.com/firojalam/assisting-fact-checking}}
\end{itemize}

\section{Related Work}
\label{sec:related_work}

Disinformation, misinformation, and ``fake news'' thrive in social media. See \cite{Lazer1094} and \cite{Vosoughi1146} for a general discussion on the science of ``fake news'' and the process of proliferation of true and false news online.
There have also been several interesting surveys, e.g.,~\citet{Shu:2017:FND:3137597.3137600} studied how information is disseminated and consumed in social media. Another survey by \citet{thorne-vlachos:2018:C18-1} took a fact-checking perspective on ``fake news'' and related problems. 
%Yet another survey \cite{Li:2016:STD:2897350.2897352} covered truth discovery in general. 

More relevant to the present work, \citet{survey:2021:ai:fact-checkers} studied how AI technology can assist professional fact-checkers, and pointed to the following research problems:
(\emph{i})~identifying claims worth fact-checking,
(\emph{ii})~detecting relevant previously fact-checked claims, (\emph{iii})~retrieving relevant evidence to fact-check a claim, and (\emph{iv})~actually verifying the claim.

The vast majority of previous work has focused on the latter problem, 
%i.e.,~claim verification, 
while the other three problems remain understudied, even though there is an awareness that they are integral steps of an end-to-end automated fact-checking pipeline \cite{vlachos2014fact,Hassan:2017:CFE:3137765.3137815}.

This situation is gradually changing, and the research community has recently started paying more attention to all four problems, in part thanks to the emergence of evaluation campaigns that feature all steps such as the CLEF CheckThat! lab.

%Here we focus on direction ({\em ii}), i.e.,~detecting relevant previously fact-checked claims, which is the least studied of the above problems. 
\citet{shaar-etal-2020-known} proposed a \emph{claim-focused} task formulation, and released two datasets: one based on PolitiFact,
% (i.e.,~claims from political debates and speeches)
 and another one based on Snopes. % (i.e.,~claims from tweets). 
They had a ranking formulation: given a claim, they asked to retrieve a ranked list of previously fact-checked claims from a given database of such claims; the database included the verified claims together with corresponding articles. 
One can argue that this formulation falls somewhere between (\emph{ii})~detecting relevant previously fact-checked claims and (\emph{iii})~retrieving relevant evidence to fact-check a claim. The same formulation was adopted at the CLEF CheckThat! lab in 2020, where the focus was on tweets, and in 2021-2022, which featured both tweets and political debates \cite{clef-checkthat-lncs:2020,clef-checkthat-en:2020,CheckThat:ECIR2020,CheckThat:ECIR2021,clef-checkthat:2021:task2,ECIR:CLEF:2022,clef-checkthat:2022:task2}.
%A similar formulation was also explored in~\cite{10.1145/3308558.3314135}.

The best systems at the CLEF CheckThat! 2021 lab used BM25 retrieval, semantic similarity using embeddings, and reranking \cite{DBLP:conf/clef/ChernyavskiyIN21,DBLP:conf/clef/MihaylovaBCHHN21,Pritzkau2021NLyticsAC}. A follow-up work used a batch softmax contrastive loss to better fine-tune BERT for the task \cite{chernyavskiy-etal-2022-batch}.

It has been further shown that it is important to match not only against the target claim, but also using the full text of the associated article that fact-checkers wrote to explain their verdict. Thus, in a follow-up work,  \citet{Claim:retrieval:context:2022} focused on modeling the context when checking an input sentence from a political debate, both on the source side and on the target side, e.g.,~by looking at neighboring sentences and using co-reference resolution.

%Other recent work on fact-checking includes \cite{si-etal-2021-topic,kazemi-etal-2021-claim,jiang-etal-2021-exploring-listwise,wan-etal-2021-dqn}.

%It was also noted that the topic of the claim and the implicit stance of the evidence towards the claim are important factors for fact-checking. To incorporate both these aspects,

\citet{sheng-etal-2021-article} proposed a re-ranker based on memory-enhanced transformers for matching (MTM) to rank fact-checked articles using key sentences selected using lexical, semantic and pattern-based similarity. \citet{si-etal-2021-topic} modeled claim-matching using topic-aware evidence reasoning and stance-aware aggregation, which model semantic interaction and topical consistency to learn latent evidence representation. 
\citet{kazemi-etal-2021-claim}  developed two datasets (one consisting of \textit{claim-like statements} and the other one using annotation of \textit{claim similarity}) 
covering four languages. 

\citet{jiang-etal-2021-exploring-listwise} used sequence-to-sequence transformer models for sentence selection and label prediction.
\citet{wan-etal-2021-dqn} proposed a deep Q-learning network, %(DQN) 
i.e., a reinforcement learning approach, which computes candidate pairs of precise evidence and their labels, and then uses post-processing to refine the candidate pairs. 

\citet{vo-lee-2020-facts} looked into multimodality. They focused on tweets that discuss images and tried to detect the corresponding verified claim by matching both the text and the image against the images in the verified claim's article. They mined their dataset from pairs of tweets and  corresponding fact-checking articles proposed by Twitter users as a response. 
\citet{AACL:2022:CrowdChecked} used a similar crowd-checking idea, and further proposed how to learn from potentially noisy data.

Finally, the task was also addressed in a reverse formulation, i.e., given a database of fact-checked claims (e.g.,~a short list of common misconceptions about COVID-19), find social media posts that make similar claims \cite{hossain-etal-2020-covidlies}.

Unlike the above work, our input is a \emph{document}, and the goal is to detect all sentences that contain a claim that can be verified by some previously fact-checked claim (from a given database).

\begin{table*}[tbh!]
    \centering
    \small
    \resizebox{1.0\textwidth}{!}{%
    % \scalebox{0.88}{
    \begin{tabular}{l  p{0.27\textwidth}  p{0.3\textwidth} p{0.17\textwidth}  l l}
        \toprule
        \bf No. & \bf \iclaim Sentence & \bf \vclaim Claim & \bf Label \& Date & \bf Stance & \bf Verdict \\
        \midrule
        % 1 & \emph{The stock market is in 22,000.} & {\bf Donald Trump:} The stock market has hit record numbers, as you know. And there has been a tremendous surge of optimism in the business world. & \textit{Mostly True}, stated on February 16, 2017 & Agree & True \\ \\
        % 2 & \emph{The stock market is in 22,000.} & {\bf Jim Webb:} The stock market has almost tripled since April of 2009. & \textit{Mostly True}, stated on March 15, 2015 & Discuss & Unknown  \\ \\
        % 1 & \emph{The reason why Democrats only talk about at the totally made up Russia story is because they have no message, no agenda, and no vision.} & \textbf{Donald Trump:} This Russia thing with Trump and Russia is a made-up story. It's an excuse by the Democrats for having lost an election that they should've won.   & \textit{Pants on Fire!}, stated on May 11, 2017 & \agree & False  \\ \\
        1 & \emph{But the Democrats, by the way, are very weak on immigration.} & \textbf{Donald Trump:} The weak illegal immigration policies of the Obama Admin. allowed bad MS 13 gangs to form in cities across U.S. We are removing them fast!   & \textit{False}, stated on April 18, 2017 & \agree & Unknown  \\ \\
        2 & \emph{ICE  we're getting MS13 out by the thousands.} & \textbf{Donald Trump:} Says of MS13 gang members, "We are getting them out of our country by the thousands."   & \textit{Mostly-False}, stated on May 15, 2018 & \agree & False  \\ \\
        3 & \emph{ICE  we're getting MS13 out by the thousands.} & \textbf{Donald Trump:} I have watched ICE liberate towns from the grasp of MS13.  & \textit{False}, stated on June 30, 2018 & \agree & Unknown  \\ \\
%        4 & \emph{While in the meantime the economy hit an all time high this morning.} & \textbf{Donald Trump:} Corporations have NEVER made as much money as they are making now.  & \textit{Half-True}, stated on August 1, 2017 & \agree & Unknown \\ \\
%        5 & \emph{But we are going to get rid of Obamacare.} & \textbf{Ed MacDougall:} Says U.S. House candidate Carlos Curbelo "opposes the repeal of Obamacare."  & \textit{Mostly-False}, stated on July 26, 2014 & \disagree & Unknown \\ \\
        4 & \emph{We have one of the highest business tax rates anywhere in the world, pushing jobs and wealth out of our country.} & \textbf{Barack Obama:} "There are so many loopholes ... our businesses pay effectively one of the lowest tax rates in the world."  & \textit{Half-True}, stated on September 26, 2008 & \disagree & Unknown \\
        \bottomrule
    \end{tabular}
    }
    \caption{Example sentences from Donald Trump's interview with \emph{Fox and Friends} on June 6, 2018.}
    \label{table:examples}
\end{table*}

\section{Task Definition}
\label{sec:task_definition}
% \todo[inline]{R1:There is no pipeline or diagram which makes it really difficult to follow the paper.}
% \fa{May be we can give a diagram to define the task?}

We define the task as follows (see also Figure \ref{fig:system_pipeline}): 

% \begin{quote}
\emph{Given an input \underline{document} and a database of previously fact-checked claims, produce a ranked list of its sentences, so that those that contain claims that can be verified by a claim from the database are ranked as high as possible. We further want the system to be able to point to the database claims that \underline{verify} a claim in an input sentence.}
% \end{quote}

%\emph{Given a transcript for a political event (i.e., \emph{document}), and a set of verified claims, rank the input sentences in the transcript, so that the input sentences that can be verified by these verified claims are on top. We further want the system to point to the actual claims that verify the selected input sentences.} \fa{In Figure \ref{fig:system_pipeline}, we present a system pipeline to have a clear understanding of the task.}

Note that we want the \iclaim sentence to be verified as true/false, and thus we want to skip matches against \vclaim claims with labels of unsure veracity such as \emph{half-true}. Note also that solving this problem requires going beyond stance, i.e.,~whether a previously fact-checked claim \agree{\emph s}/\disagree{\emph s} with the input sentence~\cite{10.1145/3308558.3314135}. In certain cases, other factors might also be important, such as, (\emph{i})~whether the two claims express the same degree of specificity, (\emph{ii})~whether they are made by the same person and during the same time period, (\emph{iii})~whether the verified claim is true/false or is of mixed factuality, etc. Table~\ref{table:examples} %in the Appendix 
shows some examples.
%of \iclaim and \vclaim claims, stance, and verdict for the \iclaim--\vclaim claim pair. 

%\fa{i think  the following sentence contradicts with the task definition, we should either use input sentence or input claim and be consistent.}Our task is to find and to rank the verified claims based on the input claim and the verdict. 

\section{Dataset}
\label{sec:dataset}

% In this section, we describe the process of creating the dataset we experiment with.

\subsection{Background}
We construct a dataset using fact-checked claims from \politifact,\footnote{\url{http://www.politifact.com/}} 
which focuses on claims by politicians.
%, e.g.,~in speeches, interviews, debates, TV shows, etc.
For each fact-checked claim, there is a factuality label and an article explaining the reason for assigning that label. PolitiFact further publishes commentaries that highlight some of the claims made in a debate or speech, with links to fact-checking articles about these claims from their website. 
These commentaries were used in previous work as a way to obtain a mapping from \iclaim sentences in a debate/speech to \vclaim claims. For example, \citet{shaar-etal-2020-known} collected 16,636 \vclaim claims and 768 \iclaim--\vclaim claim pairs from 70 debates and speeches, together with the transcript of the target event. 
For each \vclaim claim, they released  \vclaimstatement, \textit{TruthValue} \{\emph{Pants-on-Fire!}, \emph{False}, \emph{Mostly-False}, \emph{Half-True}, \emph{Mostly-True}, \emph{True}\}, \vclaimtitle and \vclaimtext.

% \begin{itemize}[noitemsep,topsep=0pt,leftmargin=10pt,labelwidth=!,labelsep=.5em]
%     % \itemsep-0.2em
%     \item \vclaimstatement: the text of the claim, which is a normalized version of the original claim, as the human fact-checkers typically reformulate it, e.g.,~to make it clearer, context-independent, and self-contained;
%     \item \textit{TruthValue}: the label of the claim \{\emph{Pants-on-Fire!}, \emph{False}, \emph{Mostly-False}, \emph{Half-True}, \emph{Mostly-True}, \emph{True}\};
% %    \footnote{Where \emph{Pants-on-Fire!} represent False \vclaim statements}
%     \item \vclaimtitle: the title of the article that discusses the claim;
%     \item \vclaimtext: the body of the article that discusses the claim.
% \end{itemize}

The above dataset has high precision, and it is suitable for their formulation of the task: given a sentence (one of the 768 ones), identify the correct claim that verifies it (from the set of 16,636 \vclaim claims). 
However, it turned out not to be suitable for our purposes due to recall issues: missing links between \iclaim sentences in the debate/speech and the set of \vclaim claims. This is because \politifact journalists were not interested in making an exhaustive list of all possible correct mappings between \iclaim sentences and \vclaim claims in their database; instead, they only pointed to some such links, which they wanted to emphasize. 

Moreover, if the debate made some claim multiple times, they would include a link for only one of these instances (or they would skip the claim altogether). Moreover, if the claims made in a sentence are verified by multiple claims in the database, they might only include a link to one of these claims (or to none).

However, we have a document-level task, where identifying sentences that can be verified using a database of fact-checked claims is our primary objective (while returning the matching claims is secondary), we need not only high precision, but also high recall for the \iclaim--\vclaim claim pairs. 

% \todo[inline]{R3: The authors do contribute a dataset in this paper. However, they do not clearly identify what has raised the need for this dataset.}

\subsection{Our Dataset}
\begin{figure}[tbh]
\centering
\includegraphics[width=0.48\textwidth]{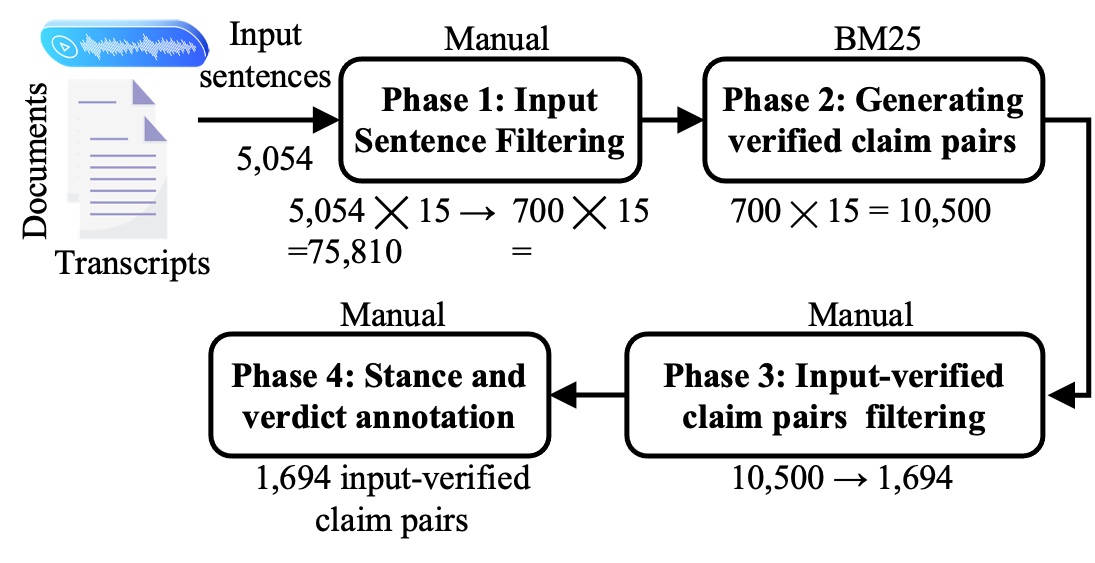}
\caption{Data preparation pipeline.}
\label{fig:data_prep_pipeline}
\end{figure}

%\fa{As the published dataset, reported in \cite{shaar-etal-2020-known}, does not serve our purposes, we decided to create our own. 
We manually checked and \textit{re-annotated} seven debates from the dataset of \citet{shaar-etal-2020-known} by linking \vclaim claims from \politifact to the \iclaim sentences in the transcript.
% \sh{\citet{shaar-etal-2020-known} has collected \iclaim--\vclaim claim pairs from 70 different debates; where a few selected \iclaim sentences were chosen from the debate and verified by a Politifact fact-checker. However, in our definition of the task we require to know all possible verified \iclaim sentences that can be verified not just the few selected by Politifact fact-checkers. Therefore, we opted to create our own dataset by manually checking and \textit{re-annotated} seven debates from the dataset of \citet{shaar-etal-2020-known} by linking \vclaim claims published on \politifact to the \iclaim sentences in the transcript.}
This includes 5,054 sentences, and ideally, we would have wanted to compare each of them against each of the 16,636 \vclaim claims, which would have resulted in a huge and very imbalanced set of \iclaim--\vclaim pairs: $5,054 \times 16,636 = 84,078,344$. 
% we first performed retrieval, and we only checked the top matches, after filtering. 
Thus, we decided to pre-filter the \iclaim sentences and the \iclaim--\vclaim claim pairs.
The process is sketched in Figure~\ref{fig:data_prep_pipeline} and described in more detail below. 
% %Below, we describe the process in detail, as also shown in Figure \ref{fig:data_prep_pipeline}.

\subsection{Phase 1: \iclaim Sentence Filtering}
Not all sentences in a speech/debate contain a verifiable factual claim, especially when uttered in a live setting. In speeches, politicians would make a claim and then would proceed to provide the numbers and the anecdotes to emphasize and to create an emotional connection with the audience. 

In our case, we only need to focus on claims. We also know that not all claims are important enough to be fact-checked. Thus, we follow \cite{DBLP:journals/corr/abs-1809-08193} 
as guidance to define which \iclaim sentences are worth fact-checking.
Based on this definition, positive examples include, but are not limited to ({\em a})~stating a definition, ({\em b})~mentioning a quantity in the present or in the past, ({\em c})~making a verifiable prediction about the future, ({\em d})~referencing laws, procedures, and rules of operation, or ({\em e})~implying correlation or causation (such correlation/causation needs to be explicit). 
Negative examples include personal opinions and preferences, among others.
%In this step, we wanted to remove as little as possible, and thus three annotators independently made judgments for check-worthiness, and we only rejected a sentence if all three chose to reject it. Eventually, we reduced the number of \emph{input sentences} to check from 5,054 to 700.
In this step, three annotators independently made judgments about the \iclaim sentences for check-worthiness (i.e.,~check-worthy \textit{vs.} not check-worthy), and we only rejected a sentence if all three annotators judged it to be not check-worthy. As a result, we reduced the number of \emph{input sentences} that need further manual checking from 5,054 to 700.

\subsection{Phase 2: Generating \iclaim--\vclaim Pairs}
Next, we indexed the \vclaim claims and we queried with the \iclaim sentence using BM25 to retrieve 15 \vclaim claims per \iclaim sentence. As a result, we managed to reduce the number of \emph{pairs} to check from $700 \times 16,636=11,645,200$ to just $700 \times 15 = 10,500$.

% to $700 \times 15 = 10,500$. \fa{The reason to choose top 15 is to reduce the manual filtering efforts as 700 $\times$16636=11,645,200 is still a large number to manually check and filter them.}
%  \fa{can we motivate why we choose 15? 700 $\times$16636=11,645,200, to still reduce such number? top 15 shows best match?}

% \paragraph{Phase 3: \vclaim Claim Filtering}
\subsection{Phase 3: \iclaim--\vclaim Pairs Filtering}
Then, we manually went through the 10,500 \iclaim--\vclaim pairs, and we filtered out the ones that were incorrectly retrieved by the BM25 algorithm. Again, we were aiming for high recall, and thus we only rejected a pair if all three out of the three annotators independently proposed to reject it. As a result, the final number of \emph{pairs} to check git reduced to just 1,694.

\begin{table*}[tbh!]
\centering
\resizebox{1.0\textwidth}{!}{%
% \scalebox{0.8}{
% \setlength{\tabcolsep}{2.0pt}
\begin{tabular}{@{}llrrrrrrr@{}}
\toprule
\multicolumn{1}{c}{\textbf{Date}} & \multicolumn{1}{c}{\textbf{Event}} & \multicolumn{1}{c}{\textbf{\# Topic}} & \multicolumn{1}{c}{\textbf{Sent.}} & \multicolumn{1}{c}{\textbf{Sent.-Var. Pairs}} & \multicolumn{1}{c}{\textbf{\# Stance-Input}} & \multicolumn{1}{c}{\textbf{\# Stance-pairs}} & \multicolumn{1}{c}{\textbf{\# Verdict-Input}} & \multicolumn{1}{c}{\textbf{\# Verdict-pairs}} \\ \midrule
2017-08-03 & Rally Speech & 3-4 & 291 & 4,365 & 34 & 62 & 20 & 32 \\
2017-08-22 & Rally Speech & 5+ & 792 & 11,880 & 50 & 116 & 23 & 40 \\
2018-04-26 & Interview & 5+ & 597 & 8,955 & 28 & 52 & 17 & 32 \\
2018-05-25 & Naval Grad. Speech & 1-2 & 279 & 4,185 & 14 & 19 & 4 & 5 \\
2018-06-12 & North Korea Summit Speech & 1-2 & 1,245 & 18,675 & 29 & 45 & 15 & 15 \\
2018-06-15 & Interview & 3-4 & 814 & 12,210 & 24 & 36 & 11 & 17 \\
2018-06-28 & Rally Speech & 5+ & 1,036 & 15,540 & 49 & 82 & 35 & 57 \\ \midrule
\textbf{Total} &  & \multicolumn{1}{l}{} & 5,054 & 75,810 & 228 & 412 & 125 & 198 \\ \bottomrule
\end{tabular}%
}

\caption{\textbf{Statistics about our dataset:} number of sentences in each transcript, and distribution of clear stance (\agree + \disagree) and clear verdict (true + false) labels. The number of topics was manually decided by looking at the keywords detected in each transcript. \emph{Sent.} is the number of input sentences, and \emph{Sent.-Var. Pairs} is the number of input sentences with top-15 verified claim pairs.}
\label{table:transcript-stance-stats}
\end{table*}

\begin{table}[tbh]
\centering
\scalebox{0.8}{
\begin{tabular}{@{}lrr@{}}
\toprule
\multicolumn{1}{c}{\textbf{Politifact Truth   Value}} & \multicolumn{1}{c}{\textbf{True/False}} & \multicolumn{1}{c}{\textbf{Unknown}} \\ \midrule
Pants on Fire! & 24 & 191 \\
FALSE & 76 & 382 \\
Mostly--False & 44 & 312 \\
Half--True & 2 & 260 \\
Mostly--True & 42 & 227 \\
TRUE & 11 & 85 \\ \bottomrule
\end{tabular}%
}
\caption{\textbf{Distribution of the labels:} \iclaim--\vclaim pairs with a true/false verdict vs. the \emph{TruthValue} for \vclaim claim from \politifact.}
\label{table:distribution}
\end{table}

\subsection{Phase 4: Stance and Verdict Annotation}
As in the previous phase, three annotators manually annotated the 1,694 \iclaim--\vclaim pairs with stance and verdict 
labels using the following label inventory:

\begin{itemize}
%[noitemsep,topsep=0pt,leftmargin=10pt,labelwidth=!,labelsep=.5em]
%[leftmargin=*,noitemsep]
    \item \textbf{stance}: \agree, \disagree, \emph{unrelated}, \emph{not--claim};
    \item \textbf{verdict}: \emph{true}, \emph{false}, \emph{unknown}, \emph{not--claim}.
\end{itemize}

The label for \textbf{stance} is \agree if the \vclaim claim agrees with the \iclaim claim, \disagree if it opposes it, and \emph{unrelated} if there is no \agree/\disagree relation (this includes truly unrelated claims or related but without agreement/disagreement, e.g.,~discussing the same topic). 

The \textbf{verdict} is \emph{true}/\emph{false} if the \iclaim sentence makes a claim whose veracity can be determined to be \emph{true}/\emph{false} based on the paired \vclaim claim and its veracity label; it is \emph{unknown} otherwise. The veracity can be unknown for various reasons, e.g.,~(\emph{i})~the \vclaim claim states something (a bit) different,
%due to the \vclaim claim, which claims something (a bit) different, 
(\emph{ii})~the two claims are about different events,
(\emph{iii})~the veracity label of the \vclaim claim is ambiguous.
We only need the verdict annotation to determine whether the \iclaim sentence is verifiable;
yet, we use the stance to construct suitable \iclaim--\vclaim claim pairs.

\subsection{Final Dataset} Our final dataset consists of 5,054 \iclaim sentences, and 75,810 \iclaim--\vclaim claim pairs. This includes 125 \iclaim sentences that can be verified using a database of 16,663 fact-checked claims, and 198 \iclaim--\vclaim claim pairs where the \vclaim claim can verify the \iclaim sentence (as some \iclaim sentences can be verified by more than one \vclaim claim). 
Table~\ref{table:transcript-stance-stats} reports some statistics about each transcript, and it also shows overall statistics (in the last row).

\subsection{Annotation and Annotators' Agreement} 
Note that each \iclaim--\vclaim claim pair was annotated by three annotators: one male and two females, with BSc and PhD degrees. The disagreements were resolved by majority voting, and, if this was not possible, in a discussion with additional consolidators.
We measured the inter-annotator agreement on phase 4 (phases 1 and 3 aimed for high recall rather than agreement).
%To understand the reliability of the annotation, we computed inter-annotator agreement. 
% In table \ref{table:annotator-agreement}, 
We obtained a Fleiss Kappa ($\kappa$) of 0.416 for stance and of 0.420 for the verdict, both corresponding to moderate level of agreement.

\section{Evaluation Measures}
\label{sec:eval_measures}
%\todo[]{WWW comments are mainly on evaluation, please check: https://docs.google.com/document/d/1b-uTxBREFFxZM4gADx5a0z8bhIzaoFjWdS5fnE5gYpU/edit?usp=sharing, any suggestion how to improve?}
Given a document, the goal is to rank its sentences, so that those that can be verified (i.e.,~with a true/false verdict; \emph{Verdict-Input} in Table~\ref{table:transcript-stance-stats}) are ranked as high as possible, and also to provide a relevant \vclaim claim (i.e.,~one that could justify the verdict; \emph{Verdict-pairs} in Table~\ref{table:transcript-stance-stats}).
This is a (double) ranking task, and thus we use ranking evaluation measures based on Mean Average Precision (MAP).
First, let us recall the standard AP:
\begin{equation}
    AP = \frac{\sum_{k=1}^{n}{P_1(k) \times rel(k)}}{{rel. sentences}}, 
\end{equation}
where $P_1(k)$ is the precision at a cut-off $k$ in the list, $rel(k)$ is 1 if the $k$-th ranked sentence is relevant (i.e.,~has either a true or a false verdict), and \textit{rel.~sentences} is the number of \iclaim sentences that can be verified in the transcript.

We define more strict AP measures, $AP_{H}^r$, $AP_0^r$, and $AP_{0.5}^r$, which only give credit for an \iclaim sentence with a known verdict, if also a corresponding \vclaim claim is correctly identified:

\begin{equation}
    AP_{H}^r = \frac{\sum_{k=1}^{n}{P_1^r(k) \times rel_H^r(k)}}{{rel. sentences}}
\end{equation}

\noindent where $rel_H^r(k)$ is 1 if the $k$-th ranked \iclaim sentence is relevant and at least one relevant \vclaim claim was retrieved in the top-$r$ \vclaim claim list.

\begin{equation}
    AP_0^r = \frac{\sum_{k=1}^{n}{P_0^r(k) \times rel(k)}}{{rel.\ sentences}}
\end{equation}

\begin{equation}
    AP_{0.5}^r = \frac{\sum_{k=1}^{n}{P_{0.5}^r(k) \times rel(k)}}{{rel.\ sentences}}
\end{equation}

\noindent where $P_{m}^r(k)$, is precision at cut-off $k$, so that it increments by $m$, if {\bf none} of the relevant \vclaim claim was retrieved in the top-$r$ \vclaim claim list; otherwise, it increments by 1.

Note that the simple $AP$ can also be represented as $AP_{1}^{r}$, as it increments by 1 regardless of whether a relevant \vclaim claim is in the top-$r$ of the list of \vclaim claims.

% \todo[inline] {Don't we increment by $m$ only when the \iclaim sentence can be checked into the database but there is no relevant \vclaim claim in the top-$r$ list?}

We compute $MAP$, $MAP_{H}^r$, $MAP_0^r$, and $MAP_{0.5}^r$ by averaging $AP$, $AP_{H}^r$, $AP_0^r$, and $AP_{0.5}^r$, respectively, over the test transcripts.

We also compute $MAP_{inner}$ by averaging the $AP_{inner}$ on the \vclaim claims: %That is, 
we compute $AP_{inner}$ for a given \iclaim sentence, by scoring the rankings of the retrieved \vclaim claims\, as in the task presented in \citep{shaar-etal-2020-known}.

\section{Model}
\label{sec:models}

% \todo[inline]{R2: - the text in Model section uses "Input claim" and "input sentence" interchangeably, which is not correct; not all sentences of the debate are claims.}
The task we are trying to solve has two subtasks. The \textit{first} sorts the \iclaim sentences in the transcript in a way, so that the \iclaim sentences that can be verified using the database are on top. The \textit{second} one consists of retrieving a list of matching \vclaim claims for a given \iclaim sentence. While we show experiments for both subtasks, our main focus is on solving the first one.

\subsection{\iclaim--\vclaim Pair Representation}
In order to rank the \iclaim sentences from the transcript, we need to find ways to represent them, so that we would have information about whether the database of \vclaim claims can indeed verify some claim from the \iclaim sentence. 

To do that, we propose to compute multiple similarity measures between all possible \iclaim--\vclaim pairs, where we can match the \iclaim sentence against the \vclaimstatement, the \vclaimtitle, and the \vclaimtext of the verified claims' fact-checking article in \politifact.

%We compute the following similarity scores between the all \iclaim--\vclaim pairs for a given \iclaim sentence, 

\begin{itemize}
%[noitemsep,topsep=0pt,leftmargin=10pt,labelwidth=!,labelsep=.5em]
%[leftmargin=*,noitemsep]
    \item \textbf{BM25~\cite{Robertson:2009:PRF:1704809.1704810}:} These are BM25 scores when matching the \iclaim sentence against the \vclaimstatement, the \vclaimtitle, and the \vclaimtext, respectively (3 features);
    % \todo{does it mean 3 scores?}
    \item \textbf{NLI Score~\cite{nie-etal-2020-adversarial}:} These are posterior probabilities for NLI over the labels \{\emph{entailment}, \emph{neutral}, \emph{contradiction}\} between the \iclaim sentence and the {\vclaimstatement} (3 features);
    \item \textbf{BERTScore~\cite{bert-score}:} F1 score from the BERTScore similarity scores between the \iclaim sentence and the {\vclaimstatement} (1 feature);
    \item \textbf{Sentence-BERT (SBERT)~\cite{reimers-gurevych-2019-sentence}:} Cosine similarity for sentence-BERT-large embedding of the \iclaim sentence compared to the embedding for the {\vclaimstatement}, the \vclaimtitle, and the \vclaimtext. Since the \vclaimtext is longer, we obtain the cosine similarity between the \iclaim sentence vs. each sentence from the \vclaimtext, and we only keep the four highest scores (6 features);
    \item \textbf{SimCSE~\cite{gao2021simcse}:} Similarly to SBERT, we compute the cosine similarity between the SimCSE embeddings of the \iclaim sentence against the {\vclaimstatement}, the \vclaimtitle, and the \vclaimtext. Again, we use the top-4 scores when matching against the \vclaimtext sentences (6 features: 1 from the \vclaimstatement + 1 from the \vclaimtitle + 4 from the \vclaimtext). 
\end{itemize}

% \begin{table}[htp]
%     \centering
%     \small
%     \resizebox{.82\linewidth}{!}{%
%     \begin{tabular}{p{0.18\textwidth} l*{4}{r}}
%         \toprule
%          & \multicolumn{2}{l}{\textbf{True} or \textbf{False}} & \multicolumn{2}{l}{\bf Unknown}\\
%         \midrule
%         \bf Politifact TruthValue & \# & \% & \# & \% \\
%         \midrule
%         Pants on Fire! & 24 & 11\% & 191 & 89\%\\
%         False & 76 & 17\% & 382 & 83\%\\
%         Mostly--False & 44 & 12\% & 312 & 82\%\\
%         Half--True & 2 & 1\% & 260 & 99\%\\
%         Mostly--True & 42 & 16\% & 227 & 84\%\\
%         True & 11 & 11\% & 85 & 89\%\\
%         \bottomrule
%     \end{tabular}
%     }
%     \caption{Distribution of the \iclaim--\vclaim pairs with a true/false verdict vs. the \emph{TruthValue} of the \vclaim claim from \politifact.}
%     \label{table:distribution}
% \end{table}

% \begin{table}[htp]
% \centering
% \scalebox{0.8}{
% \begin{tabular}{@{}lrr@{}}
% \toprule
% \multicolumn{1}{c}{\textbf{Politifact Truth   Value}} & \multicolumn{1}{c}{\textbf{True/False}} & \multicolumn{1}{c}{\textbf{Unknown}} \\ \midrule
% Pants on Fire! & 24 & 191 \\
% FALSE & 76 & 382 \\
% Mostly--False & 44 & 312 \\
% Half--True & 2 & 260 \\
% Mostly--True & 42 & 227 \\
% TRUE & 11 & 85 \\ \bottomrule
% \end{tabular}%
% }
% \caption{\textbf{Distribution:} \iclaim--\vclaim pairs with a true/false verdict vs. the \emph{TruthValue} for \vclaim claim from \politifact.}
% \label{table:distribution}
% \end{table}

% \input{sections/result-table}

\subsection{Single-Score Baselines}
Each of the above scores, e.g.,~SBERT, can be calculated for each \iclaim--\vclaim claim pair. For a given \iclaim sentence, this makes 16,663 scores (one for each \vclaim from the database), and as a baseline, we assign to the \iclaim sentence the maximum over these scores. Then, we sort the sentences of the input document based on these scores, and we evaluate the resulting ranking.

\begin{table}[t]
    \centering
    \small
    \resizebox{1.0\linewidth}{!}{%
    \begin{tabular}{lc}
        \toprule
        \bf Experiment & \bf MAP$_{inner}$ \\
        \midrule 
        BERTScore (F1) on \vclaimstatement & 0.638 \\
		NLI (Entl) on \vclaimstatement & 0.574 \\
		NLI (Neut) on \vclaimstatement & 0.112 \\
		NLI (Contr) on \vclaimstatement & 0.025 \\
		NLI (Entl+Contr) on \vclaimstatement & 0.553 \\
		SimCSE on \vclaimtitle & 0.220 \\
		SimCSE on \vclaimstatement & 0.451 \\
		SimCSE on \vclaimtext & 0.576 \\
		SBERT on \vclaimtitle & 0.165 \\
		SBERT on \vclaimstatement & 0.531 \\
		SBERT on \vclaimtext &  0.649 \\
		BM25 on \vclaimstatement & 0.316 \\
		BM25 on \vclaimtext & \bf 0.892 \\
		BM25 on \vclaimtitle & 0.145 \\
        \bottomrule
    \end{tabular}
    }
    \caption{\textbf{Preliminary \vclaim Claim retrieval experiments} on the annotations obtained from the \politifact dataset and the manually annotated pairs with \agree or \disagree stance.}
    \label{table:inner-map-experiments}
\end{table}

\begin{table*}[h]
    \centering
    \resizebox{0.8\textwidth}{!}{%
    % \scalebox{0.75}{
    \begin{tabular}{p{0.45\textwidth}l*{8}{c}}
        \toprule
        \bf Experiment & \bf MAP & \bf MAP$_0^1$ & \bf MAP$_0^3$ & \bf MAP$_{0.5}^1$ & \bf MAP$_{0.5}^3$ & \bf MAP$_{H}^1$ & \bf MAP$_{H}^3$ \\
        \midrule 
        \multicolumn{8}{c}{\bf Baselines: Single Scores}\\
        \midrule
        BERTScore (F1) on \vclaimstatement & 0.076 & 0.046 & 0.050 & 0.061 & 0.063 & 0.034 & 0.038 \\
		NLI (Entl) on \vclaimstatement  & 0.035 & 0.025 & 0.029 & 0.030 & 0.032 & 0.017 & 0.023 \\
		NLI (Neut) on \vclaimstatement  & 0.036 & 0.001 & 0.003 & 0.019 & 0.020 & 0.000 & 0.001 \\
		NLI (Contr) on \vclaimstatement  & 0.051 & 0.001 & 0.001 & 0.026 & 0.026 & 0.000 & 0.000 \\
		NLI (Entl+Contr) on \vclaimstatement& 0.041 & 0.005 & 0.007 & 0.023 & 0.024 & 0.002 & 0.003 \\
		SimCSE on \vclaimstatement       & 0.287 & 0.249 & 0.259 & 0.268 & 0.273 & \bf 0.208 & 0.223 \\
		SimCSE on \vclaimtitle           & 0.242 & 0.144 & 0.213 & 0.193 & 0.227 & 0.093 & 0.172 \\
		SimCSE on \vclaimtext            & 0.068 & 0.041 & 0.048 & 0.055 & 0.058 & 0.025 & 0.034 \\
		SBERT on \vclaimstatement        & \bf 0.303 & \bf 0.245 & \bf 0.284 & \bf 0.274 & \bf 0.294 & 0.203 & \bf 0.251 \\
		SBERT on \vclaimtitle            & 0.117 & 0.044 & 0.082 & 0.080 & 0.099 & 0.019 & 0.060 \\
		SBERT on \vclaimtext             & 0.033 & 0.016 & 0.021 & 0.025 & 0.027 & 0.008 & 0.012 \\
		BM25 on \vclaimstatement          & 0.146 & 0.107 & 0.122 & 0.127 & 0.134 & 0.086 & 0.100 \\
		BM25 on \vclaimtitle             & 0.084 & 0.047 & 0.049 & 0.066 & 0.067 & 0.031 & 0.034 \\
		BM25 on \vclaimtext              & 0.155 & 0.130 & 0.144 & 0.143 & 0.150 & 0.107 & 0.132 \\
        \midrule 
        \multicolumn{8}{c}{\bf RankSVM for Retrieved \vclaim Claims (using BM25 on \vclaimtext)}\\
        \midrule
        Top-1   & 0.382 & 0.357 & 0.373 & 0.369 & 0.378 & 0.310 & 0.352 \\
		Top-3   & 0.345 & 0.318 & 0.336 & 0.332 & 0.341 & 0.278 & 0.319 \\
		Top-5   & 0.362 & 0.335 & 0.353 & 0.349 & 0.357 & 0.292 & 0.335 \\
		Top-10  & \bf 0.404 & \bf 0.364 & \bf 0.391 & \bf 0.384 & \bf 0.398 & \bf 0.313 & \bf 0.368 \\
		Top-20  & 0.400 & 0.346 & 0.377 & 0.373 & 0.388 & 0.291 & 0.352 \\
		Top-30  & 0.357 & 0.310 & 0.339 & 0.333 & 0.348 & 0.260 & 0.318 \\
		\midrule 
		\multicolumn{8}{c}{\bf RankSVM--Max}\\
		\midrule
		Top-1  & 0.411 & 0.299 & 0.390 & 0.355 & 0.401 & 0.253 & 0.364 \\
		Top-3  & 0.449 & 0.328 & 0.429 & 0.389 & 0.439 & 0.273 & 0.400 \\
		Top-5  & 0.482 & 0.349 & 0.464 & 0.416 & 0.473 & 0.291 & 0.436 \\
		Top-10 & \bf 0.491 & \bf 0.394 & \bf 0.473 & \bf 0.443 & \bf 0.482 & \bf 0.320 & \bf 0.445 \\
		Top-20 & 0.488 & 0.381 & 0.470 & 0.434 & 0.479 & 0.310 & 0.439 \\
		Top-30 & 0.486 & 0.377 & 0.468 & 0.432 & 0.477 & 0.304 & 0.435 \\
		\midrule 
		\multicolumn{8}{c}{\bf RankSVM--Max with Skipping Half-True \vclaim claims}\\
		\midrule
		Top-1  & 0.467 & 0.353 & 0.442 & 0.410 & 0.455 & 0.287 & 0.417\\
		Top-3  & 0.507 & 0.370 & 0.485 & 0.438 & 0.496 & 0.306 & 0.454\\
		Top-5  & \underline{\bf 0.522} & 0.379 & \underline{\bf 0.501} & 0.451 & \underline{\bf 0.512} & 0.316 & \underline{\bf 0.468}\\
		Top-10 & 0.515 & \underline{\bf 0.401} & 0.494 & \underline{\bf 0.458} & 0.505 & \underline{\bf 0.323} & 0.465\\
		Top-20 & 0.504 & 0.350 & 0.481 & 0.427 & 0.493 & 0.293 & 0.447\\
		Top-30 & 0.493 & 0.376 & 0.468 & 0.435 & 0.481 & 0.301 & 0.433\\
        \bottomrule
    \end{tabular}
    }
    \caption{\textbf{Verdict experiments:} Baseline and re-ranking experiments on the \politifact dataset. The results highlighted in \textbf{bold} are the best results for the particular sets of experiments. The \underline{underlined} results are the best overall.}
    \label{table:baseline-ranksvm-experiments}
\end{table*}

\begin{table*}[tbh]
    \centering
    \resizebox{0.8\textwidth}{!}{%
    % \scalebox{0.75}{
    \begin{tabular}{p{0.45\textwidth}l*{8}{c}}
        \toprule
        \bf Experiment & \bf MAP & \bf MAP$_0^1$ & \bf MAP$_0^3$ & \bf MAP$_{0.5}^1$ & \bf MAP$_{0.5}^3$ & \bf MAP$_{H}^1$ & \bf MAP$_{H}^3$ \\
        \midrule
        RankSVM--Max on Top-5 with Skipping  & 0.522 & 0.379 & 0.501 & 0.451 &  0.512 & 0.316 & 0.468\\
        \midrule
        w/o BERTScore (F1) & 0.499 & 0.376 & 0.480 & 0.437 & 0.489 & 0.313 & 0.450 \\
		w/o NLI Score (E, N, C) & 0.475 & 0.330 & 0.451 & 0.402 & 0.463 & 0.279 & 0.423 \\
		w/o SimCSE & 0.511 & 0.353 & 0.486 & 0.432 & 0.499 & 0.295 & 0.454 \\
		w/o SBERT & 0.498 & 0.381 & 0.481 & 0.440 & 0.490 & 0.308 & 0.452 \\
		w/o BM25 & 0.497 & 0.343 & 0.473 & 0.420 & 0.485 & 0.287 & 0.441 \\
		w/o scores on \vclaimtitle & 0.522 & 0.369 & 0.501 & 0.445 & 0.511 & 0.308 & 0.468 \\
		w/o scores on \vclaimstatement & 0.311 & 0.242 & 0.293 & 0.276 & 0.302 & 0.198 & 0.268 \\
		w/o scores on \vclaimtext & 0.444 & 0.295 & 0.427 & 0.370 & 0.435 & 0.249 & 0.398 \\
		\bottomrule
    \end{tabular}
    }
    \caption{\textbf{Ablation experiments for the verdict} on the best model from Table~\ref{table:baseline-ranksvm-experiments}: RankSVM with Top-5 scores from all measures while skipping \textit{half-true} \vclaim claims.}
    \label{table:ablation-experiments}
\end{table*}

% \todo[inline]{R1: Calculating 16k similarity scores for each input as a baseline seems kinda strange. Can you not do with a better baseline in terms of filtering?} \fa{May be we can say why we have chosen this as a baseline?}

\subsection{Re-ranking Models}

We performed preliminary experiments looking into how the above measures work for retrieving the correct \vclaim claim for an \iclaim sentence for which there is at least one match in the \vclaim claims database. This corresponds to the sentence-level task of \cite{shaar-etal-2020-known}, but on our dataset, where we augment the matching \iclaim--\vclaim pairs from their dataset with all the \iclaim--\vclaim pairs with a stance of \agree or \disagree. The results are shown in Table~\ref{table:inner-map-experiments}. We can see that \emph{BM25 on Body} yields the best overall MAP score, which matches the observations in \cite{shaar-etal-2020-known}.

\paragraph{RankSVM for \vclaim Claim Retrieval}
Since now we know that the best \vclaim claim retriever uses \emph{BM25 on Body}, we use it to retrieve the top-$N$ \vclaim claims for the \iclaim sentence, and then we calculate the above 19 similarity measures for each candidate in this top-$N$ list. Afterwards, we concatenate the scores for these top-$N$ candidates. Thus, we create a feature vector of size $19 \times N$ for each \iclaim sentence.
For example, a top-3 experiment uses for each \iclaim sentence a feature vector of size $19 \times 3 = 57$, which represents each similarity measure based on the top-3 \vclaim claims retrieved by \emph{BM25 on Body}. Then, we train a RankSVM model using this feature representation.

\paragraph{RankSVM--Max}
Instead of concatenating the 19-dimensional vectors for the top-$N$ candidates, we take the maximum over these candidates for each feature to obtain a new 19-dimensional vector. 
%The hypothesis is that the further apart these scores are, the more confident we can be that the \iclaim sentence can be verified by the top retrieved \vclaim claim \cite{yang2019simple}.
Then, we train a RankSVM model like before.

% We turn here our attention to the distribution of the 19 best scores (one for each similarity measure above) over all \vclaim claims. Our
% %The next experiments sets confidence in the \iclaim sentence by looking at the top-$N$ scores and how much they differ form one another. The 
% hypothesis is that the more such scores are far apart from each other, the more confident we can be that the \iclaim sentence can be verified by the top retrieved \vclaim \cite{yang2019simple}.  %Therefore, we represent a sentence by using the top-$N$ scores of each metric regardless of which \vclaim claim they were obtained from. For example, it could be that the top score from \emph{BM25 on \vclaimstatement} is different than the \emph{BERTScore on \vclaimstatement}. 

\paragraph{RankSVM--Max with Skipping}
%While further analysing the annotations done, we can conclude from 
Table~\ref{table:distribution} shows that almost all \iclaim--\vclaim pairs for which the \emph{TruthValue} of the \vclaim claim is Half--True eventually result in an \iclaim sentence for which we cannot determine an actual verdict. This is to be expected as, if we cannot trust the veracity of the \vclaim claim, then even if the statement matches the \iclaim sentence, we cannot determine its veracity. 
Thus, we further experiment with a variant of the \textbf{RankSVM--Max} model that skips any scores that belong to a Half--True \vclaim claim. 

\section{Experiments and Evaluation}
\label{sec:exp_eval}

% We performed a 7-fold cross-validation, where we used 6 out of the 7 transcripts for training and the remaining one for testing. We started by computing the 19 similarity measures described in Section~\ref{sec:models}. Then, we used these representations to test the baselines and to train pairwise learning-to-rank models. The results are shown in Table~\ref{table:baseline-ranksvm-experiments}.
We performed a 7-fold cross-validation, where we used 6 out of the 7 transcripts for training and the remaining one for testing.
We first computed 19 similarity measures and then used them to test the baselines and to train pairwise learning-to-rank models. The results are shown in Table~\ref{table:baseline-ranksvm-experiments}.

\subsection{Baselines}

First, we discuss the results for the baseline experiments.

%For each similarity measure, we compute the similarity score between the \iclaim sentences from the test transcript and all the \vclaim claims in the database. We later use the max score between the \iclaim sentence and the \vclaim claims as the prediction score. 
We can see in Table \ref{table:baseline-ranksvm-experiments} that Sentence-BERT and SimCSE, when computed on the \vclaim claims, perform best.
An interesting observation can be made by comparing Table~\ref{table:inner-map-experiments} and Table~\ref{table:baseline-ranksvm-experiments}. In Table \ref{table:inner-map-experiments}, we see that the best \vclaim claim retriever uses BM25 on \vclaimtext; however, Table~\ref{table:baseline-ranksvm-experiments} shows poor results when we try to use BM25 to rerank \iclaim sentences. 

Moreover, while in Table~\ref{table:baseline-ranksvm-experiments} the best-performing model uses SBERT calculated on \vclaimstatement, 
Table~\ref{table:inner-map-experiments} shows that the \vclaim retriever using that model performs quite poorly.
Our investigation showed that this is because SBERT tends to yield high scores for \vclaim claims, even when there is no relevant \vclaim claim. Thus, it can be a matter of calibration.

\subsection{RankSVM for \vclaim Claims Retrieval}
We trained a RankSVM on the 19 similarity measures computed for the top-$N$ retrieved \vclaim claims, according to BM25, the best system on \vclaimtext. We can see from Table~\ref{table:baseline-ranksvm-experiments} that using the RankSVM on the 19 measures improves the scores by up to 10 MAP points absolute. Moreover, the best model achieves a MAP score of 0.404. 

\subsection{RankSVM--Max}
Using max-pooling instead of BM25-retrieved \vclaim claims yields huge improvements in MAP: from 0.404 to 0.491 using RankSVM on the top-10 scores from the 19 metrics. 

A sizable improvement can be observed when we consider MAP$_0^3$, MAP$_{0.5}^3$ and MAP$_{H}^3$ from RankSVM for \vclaim claims retrieval. % since the overall reranker performs better. 
Note that, since there is a max over each measure independently, we no longer have a unified \vclaim suggestion, which is required to compute MAP$_0$, MAP$_{0.5}$, and MAP$_H$. Thus, to compute them, we use the best \vclaim claim retriever from Table~\ref{table:inner-map-experiments}, i.e.,~BM25 on \vclaimtext.

\subsection{RankSVM--Max with Skipping}
The highest MAP score, 0.522, is achieved by the RankSVM that uses the top-5 scores from each measure while skipping the Half--True \vclaim claim scores. We can also conclude by looking at the other variants of the MAP score, e.g., MAP$_H$, that we can identify the \iclaim sentences that need to be fact-checked and detect the correct \vclaim claims in the top-3 ranks.

\subsection{Ablation Experiments}
We performed an ablation study for the best-performing model in Table~\ref{table:baseline-ranksvm-experiments}, by removing one feature at a time. 
We also excluded all scores based on \vclaimtitle, \vclaimstatement, and \vclaimtext. The results are shown in Table~\ref{table:ablation-experiments}. 

We can see that the largest drops, and therefore the most important features, are the \vclaimstatement and \vclaimtext scores, whereas without \vclaimtitle scores the model performs almost identically to the original. 
We also notice that although the NLI Score did not perform very well by itself (see the baselines in Table~\ref{table:baseline-ranksvm-experiments}), it yields a significant drop, from 0.522 to 0.475 MAP points, when it is removed, which shows that it is indeed quite important.

\section{Conclusion and Future Work}
\label{sec:conclusion}

We introduced a new challenging real-world task formulation to assist fact-checkers, journalists, media, and regulatory authorities in finding which claims in a long document have been previously fact-checked. Given an input document, we aim to detect all sentences containing a claim that can be verified by some previously fact-checked claims (from a given database of previously fact-checked claims).
We developed a new dataset for this task formulation, consisting of seven debates, 5,054 sentences, 16,636 target verified claims to match against, and 75,810 manually annotated sentence--verified claim pairs.

We further defined new evaluation measures (variants of MAP), which are better tailored for our task setup.
We addressed the problem using learning-to-rank, and we demonstrated sizable performance gains over strong baselines.
We offered analysis and discussion, which can facilitate future research, and we released our data and code.

In future work, we plan to focus more on detecting the matching claims, which was our second objective here. We also plan to explore other transformer architectures and novel ranking approaches such as multi-stage document ranking using monoBERT and duoBERT \cite{yates2021pretrained}.

\section{Limitations}
\label{sec:limitations}
We have developed a dataset and proposed and evaluated a model using data from PolitiFact, which consists of political statements. We have not evaluated our approach on other topics, e.g., factual claims appearing on social media, which is out of the scope of the present work.

\section*{Ethics and Broader Impact}

\paragraph{Biases}

We note that there might be some biases in the data we use, as well as in some judgments for claim matching. These biases, in turn, will likely be exacerbated by the unsupervised models trained on them. This is beyond our control, as the potential biases in pre-trained large-scale transformers such as BERT and RoBERTa, which we use in our experiments.

\paragraph{Intended Use and Misuse Potential}

Our models can make it possible to put politicians on the spot in real time, e.g., during an interview or a political debate, by providing journalists with tools to do trustable fact-checking in real time. They can also save a lot of time to fact-checkers for unnecessary double-checking something that was already fact-checked. However, these models could also be misused by malicious actors. We, therefore, ask researchers to exercise caution.

\paragraph{Environmental Impact}
We would like to warn that the use of large-scale Transformers requires a lot of computations and the use of GPUs/TPUs for training, which contributes to global warming \cite{strubell-etal-2019-energy}. This is a bit less of an issue in our case, as we do not train such models from scratch; rather, we fine-tune them on relatively small datasets. Moreover, running on a CPU for inference, once the model is fine-tuned, is perfectly feasible, contributes much less to global warming.

\bibliographystyle{acl_natbib}
\bibliography{bib/custom}

% \newpage
% \clearpage
% \section*{Appendix}
% \label{sec:appendix}
% \appendix
% \input{sections/appendix}

% \bibliographystyle{acl_natbib}

\end{document}